\documentclass[twoside,11pt]{article}
\usepackage{jmlr2e_arxiv}

\usepackage[utf8]{inputenc} % allow utf-8 input
\usepackage[T1]{fontenc}    % use 8-bit T1 fonts
\usepackage{url}            % simple URL typesetting
\usepackage{amsfonts}       % blackboard math symbols
\usepackage{nicefrac}       % compact symbols for 1/2, etc.
\usepackage{microtype}      % microtypography

\usepackage{graphicx}
\usepackage{subfigure}
\usepackage{booktabs} % for professional tables

\usepackage{hyperref}
\usepackage[]{graphics}
\usepackage{amsmath, amssymb, amsthm}
\usepackage{epsfig}
\usepackage{verbatim}
\usepackage{setspace}
\usepackage{natbib}
\usepackage{algorithm}
\usepackage{algorithmic}

\usepackage{xcolor}

\newcommand{\R}{\mathbb{R}}
\newcommand{\Nc}{\mathcal{N}}

\newcommand{\Lc}{\mathcal{L}}
\newcommand{\Dc}{\mathcal{D}}
\newcommand{\Xc}{\mathcal{X}}

\newcommand{\Rc}{\mathcal{R}}
\newcommand{\Mc}{\mathcal{M}}
\newcommand{\Ac}{\mathcal{A}}
\newcommand{\Sc}{\mathcal{S}}
\newcommand{\Pc}{\mathcal{P}}

\newcommand{\E}{\mathbb{E}}

\newtheorem{assumption}{Assumption}

\jmlrheading{1}{2000}{1-48}{4/00}{10/00}{dwaracherla21}{Vikranth Dwaracherla and Benjamin Van Roy}

\ShortHeadings{Langevin DQN}{Dwaracherla, Van Roy}
\firstpageno{1}

\begin{document}
\title{Langevin DQN}
\author{\name Vikranth Dwaracherla \email vikranth@stanford.edu \\
\addr Stanford University
\AND 
\name Benjamin Van Roy \email bvr@stanford.edu \\
\addr Stanford University
}
\editor{}
\maketitle

%===============================================================

\begin{abstract}

Algorithms that tackle deep exploration -- an important challenge in reinforcement learning -- have relied on epistemic uncertainty representation through ensembles or other hypermodels, exploration bonuses, or visitation count distributions.  An open question is whether deep exploration can be achieved by an incremental reinforcement learning algorithm that tracks a single point estimate, without additional complexity required to account for epistemic uncertainty.  We answer this question in the affirmative. In particular, we develop Langevin DQN, a variation of DQN that differs only in perturbing parameter updates with Gaussian noise and demonstrate through a computational study that the presented algorithm achieves deep exploration.  We also offer some intuition to how Langevin DQN achieves deep exploration. In addition, we present a modification of the Langevin DQN algorithm to improve the computational efficiency.

\end{abstract}

%===============================================================

% Introduction
%!TEX root = main.tex

\section{Introduction}

In reinforcement learning (RL), intelligent exploration relies on decisions that are driven not only by expectations but also epistemic uncertainty.  Thompson Sampling (TS), for example, is a popular exploration scheme that makes decisions based on a posterior distribution over models \citep{thompson1933likelihood,russo2018tutorial}.  In its basic form, to generate decision, TS samples a model from the posterior and then selects an action that optimizes the sampled model.

Generating exact posterior samples is computationally feasible only for very simple environments, like tabular MDPs with Dirichlet priors over transition probability vectors \citep{osband2013more}.  Scaling TS to complex domains calls for approximations \citep{russo2018tutorial}. To serve this need, \citet{osband2017deep} developed randomized least-squares value iteration (RLSVI), which aims to approximately sample from the posterior over the optimal value function without explicitly representing the distribution. This algorithm randomly perturbs a prior and an accumulated dataset and fits, to this perturbed prior and data, a point estimate of the value function. The randomness induced by perturbations induces an intelligent form of exploration in actions subsequently selected to maximize the resulting value estimates.

Though RLSVI avoids maintaining an explicit posterior distribution, for each episode of operation, the algorithm produces a new point estimate based on an independently perturbed data set and prior. This requires intensive computations which do not leverage previously computed point estimates. Ensemble sampling \citep{osband2016deep,osband2017deep,lu2017ensemble} can approximate the performance of RLSVI via maintaining a set of point estimates, each updated incrementally as data accumulates. However, maintaining an ensemble of complex models is itself computationally burdensome. Moreover, for a good approximation of the posterior distribution, the ensemble size needs to grow drastically with the complexity of the posterior distribution \citep{dwaracherla2020hypermodels}.

As an alternative to maintaining an ensemble of models, one can instead learn a {\it hypermodel}, which can be used to generate approximate posterior samples, as discussed in \citep{rldm,dwaracherla2020hypermodels}. This is a promising approach, but it requires a representation that can be much more complex than a point estimate of the value function. 

An important feature of RLSVI and incremental variants that leverage ensembles or more general hypermodels is that they exhibit intelligence in the form of {\it deep exploration}.  In particular, actions are taken to resolve epistemic uncertainty not only based on immediate consequences but also on what will be observed over the subsequent time periods.  In this paper, we seek to understand whether a reinforcement learning algorithm that evolves a single point estimate of the optimal value function through incremental training can achieve deep exploration.  We propose an algorithm -- Langevin DQN -- which we believe to serve this purpose.  This algorithm synthesizes DQN \citep{mnih2013playing} and stochastic gradient Langevin iterations \citep{welling2011bayesian}.

On the surface, Langevin DQN may appear to share motivation and spirit with the ``noisy networks’’ version of DQN \citep{fortunato2018noisy}, which randomly perturbs neural network weights both when computing state-action and target values for training and in selecting actions.  But there are critical differences, as is evidenced by the fact that noisy networks do not achieve deep exploration \citep{osband2018randomized}.

Since the introduction of stochastic gradient Langevin dynamics (SGLD) \citep{welling2011bayesian}, it has been widely studied both theoretically and empirically. However, its application in sequential decision problems is at a nascent stage. \citet{mazumdar2020approximate} has studied the use of SGLD in a bandit setting with independent arms and presented optimal regret guarantees with only a constant computation cost per iteration. In contrast, our work, Langevin DQN can be applied to reinforcement learning problems and has the ability to perform deep exploration using a fixed number of update steps per episode. Through our work we demonstrate that it is possible to achieve deep exploration with a single point estimate, which is updated incrementally, without additional complexity to account for epistemic uncertainty. Moreover, Langevin DQN differs from DQN only by a Gaussian noise term in the update step.

We also provide a variant of Langevin DQN, Ensemble Langevin DQN, based on Ensemble Sampling \citep{osband2016deep, osband2017deep, lu2017ensemble}. This has improved over the capability of Boot DQN (a widely used Ensemble Sampling algorithm) to perform {\it deep exploration}. Ensemble Langevin DQN also enables us to incorporate complicated priors in Langevin DQN algorithm using prior functions \citep{osband2018randomized}.

We provide some background on Langevin Monte Carlo and episodic reinforcement learning problem in Section \ref{sec:langevin-sgd} and \ref{sec:rl-problem}, respectively. We introduce our proposed Langevin DQN algorithm in Section \ref{sec:algorithm} and show its efficacy to perform deep exploration through computational results on an environment known as {\it deep sea}. In Section \ref{sec:analysis}, we provide some intuition on why the Langevin DQN algorithm exhibits deep exploration. In Section \ref{sec:ensemble_langevin}, we present Ensemble Langevin DQN a modification to Langevin DQN to improve computational complexity while retaining its ability to perform deep exploration and demonstrate this using a wide range of computational results. Finally, we conclude in Section \ref{sec:conclusion}. Additional results and experiment specifications are provided in the appendix.

% Introducing the algorithm
%!TEX root = main.tex

\section{ Langevin SGD} \label{sec:langevin-sgd}

Langevin Monte Carlo is a Markov chain Monte-Carlo (MCMC) algorithm that can generate samples approximately from a desired probability distribution.  The algorithm represents a discrete-time approximation to the Langevin diffusion process, which takes the form
\begin{align*}
    d\overline{\theta}_t =  \nabla_{\theta =\overline{\theta}_t} \log p(\theta) dt + \sqrt{2} dB_t,
\end{align*}
where $\overline{\theta}_t$ takes values in $\Re^d$, $p$ is the probability density of interest, and $B_t$ is a standard Brownian motion in $\Re^d$.
Under suitable technical conditions, it can be shown that $p$ is the unique invariant distribution of the Langevin diffusion process (see e.g., Proposition 6.1 in \citep{pavliotis2014stochastic}).

A common Langevin Monte Carlo algorithm arises from Euler-Maruyama discretization:  
\begin{align}
    \Delta\theta_k =  \epsilon_k \nabla_{\theta = \theta_{k-1}} \log p(\theta) + \sqrt{2 \epsilon_k} z_k,
    \label{eq:LangevinMC}
\end{align}
where $\Delta\theta_k = \theta_k - \theta_{k-1}$, $\epsilon_k \in \Re_+$ is a step size, $z_k$ is independently sampled from the $d$-dimensional standard normal distribution, and $\theta_k$ represents an approximation of $\overline{\theta}_t$ with $t = \sum_{j=1}^k \epsilon_j$.  The hope is that, under technical regularity conditions and with a suitable step size sequence, for sufficiently large $k$, the marginal distribution of $\theta_k$ is close to $p$ (for example, non-asymptotic convergence rates with respect to the 1-Wasserstein distance are provided in \citep{cheng2018sharp}). 

Before we develop Langevin DQN, it is helpful to first discuss a simpler but closely related application of Langevin Monte Carlo, which we will refer to as Langevin SGD.  In particular, consider a supervised learning problem, where given a set $\Dc = \{(x_i, y_i) \}_{i=1}^T$ of data pairs, each in $\Xc \times \Re$, the goal is to fit a function $f_{\theta} : \Xc \to \R$, which is parameterized by $\theta \in \R^d$.
If we assume that $y_i = f_{\theta^*}(x_i) + w_i$ for some $\theta^* \in \Re^d$ and iid $w_i \sim \Nc(0,\sigma_w^2)$ then the negative log-likelihood function is given by
\begin{equation*}
    \Lc(\theta, \Dc)  = \sum_{i=1}^T \frac{(f_{\theta}(x_i) - y_i)^2}{2\sigma_w^2} 
\end{equation*}
Further, letting $\psi$ denote the negative log prior density of $\theta^*$, the posterior log-density of $\theta^*$ is given by
$$\log p(\theta | \Dc) = - \Lc(\theta, \Dc) - \psi(\theta).$$

Specialized to our supervised learning context, the Langevin Monte Carlo algorithm \eqref{eq:LangevinMC} becomes
\begin{align}
    \Delta\theta_k =&  \epsilon_k \nabla_{\theta = \theta_{k-1}} \log p(\theta | \Dc)  + \sqrt{2 \epsilon_k} z_k, \nonumber \\
    = & - \epsilon_k \nabla_{\theta=\theta_{k-1}}\left(\Lc(\theta,\Dc) + \psi(\theta)\right) + \sqrt{2\epsilon_k}z_k. \label{eq:langmc_post2}
\end{align}
This iteration is like gradient descent, but with an extra term that injects noise into the parameter update.  Let us refer to this as {\it Langevin gradient descent}.  While gradient descent aims to converge on a MAP estimate of $\theta^*$, Langevin gradient descent generates a stochastic sequence of parameter vectors $\theta_0, \theta_1,\theta_2, \ldots$ such that, under suitable technical conditions and for sufficiently large $k$, the marginal distribution of $\theta_k$ offers a close approximation to the posterior distribution of $\theta^*$.  

When the dataset $\Dc$ is large, processing the entire dataset in each iteration may be computationally onerous.  It is thus common to use stochastic gradient descent (SGD), which carries out each update step using a minibatch of data, rather than standard gradient descent.  Similarly, we can replace each Langevin gradient descent step \eqref{eq:langmc_post2} with one computed from a minibatch:
\begin{equation}
    \Delta\theta_k = - \epsilon_k \nabla_{\theta = \theta_{k-1}}\left( \frac{|\Dc|}{|\tilde{\Dc}|}\Lc(\theta, \tilde{\Dc}) + \psi(\theta) \right) + \sqrt{2\epsilon_k}z_k, \label{eq:sgld_0}
\end{equation} 
where $\tilde{\Dc} \subseteq \Dc$ is a fixed size minibatch, for which data points are sampled uniformly at random from $\Dc$ with replacement. We will refer to this algorithm as {\it Langevin SGD}.

Note that the first term in Equation \ref{eq:sgld_0} is an unbiased estimator of the first term in Equation \ref{eq:langmc_post2}. \citet{welling2011bayesian} argue that, for an appropriately chosen step size sequence, the marginal distribution of $\theta_k$ converges to the posterior distribution in a sense that can be made precise.

To introduce notation that simplifies development of Langevin DQN, let $\alpha_k = \epsilon_k|\Dc|$ so that we can rewrite Langevin SGD \eqref{eq:sgld_0} as
\begin{align}
    \Delta\theta_k  = - \alpha_k \nabla_{\theta = \theta_{k-1}}& \left(\frac{\Lc(\theta,\tilde{\Dc})}{|\tilde{\Dc}|} + \frac{\psi(\theta)}{|\Dc|}\right) + \sqrt{\frac{2\alpha_k}{|\Dc|}}z_k. \label{eq:sgld}
\end{align}

\section{Reinforcement Learning} \label{sec:rl-problem}

Langevin DQN synthesizes Langevin SGD and DQN to address RL.  In this paper, we restrict attention to an episodic RL setting in which an agent interacts with an unknown environment over episodes, aiming to maximize accumulated rewards. We model the environment as a Markov Decision Process (MDP) characterized by a quintuple $\Mc = (\Sc, \Ac, \Rc, \Pc, \rho)$. Here, $\Sc$ is a finite state space, $\Ac$ is a finite action space, $\Rc$ is a reward model, $\Pc$ is a transition model, and $\rho$ is an initial state distribution. Given a state $s \in \Sc$, action $a \in \Ac$, and next state $s’ \in \Sc$, $\Rc_{s,a,s’}(\cdot)$ denotes a distribution of the possible rewards the agent can experience when transitioning from $s$ to $s’$ upon taking action $a$. Similarly, $\Pc_{s,a}(s’)$ is the conditional probability that the state transitions to $s’$ from state $s$ upon taking action $a$, while $1 - \sum_{s’ \in \Sc}\Pc_{s,a}(s’)$ is the probability that the episode terminates. We denote the sequence of observations made by the agent in episode $l$ by $(s_0^l, a_0^l, r_0^l, s_1^l, \dots s_{\tau^l-1}^l,  a_{\tau^l-1}^l, r_{\tau^l-1}^l, s_{\tau^l}^l )$, where $s^l_h$ is the state of the environment observed by the agent at time $h$ and episode $l$, $r^l_h$ is the reward observed by the agent on taking an action $a^l_h$, and $\tau^l$ denotes the time at which episode terminates at state $s^l_{\tau^l}$.

A (stationary stochastic) policy $\pi : \Sc \times \Ac \to [0,1]$ assigns a probability mass function over actions to each state. In particular, we denote by $\pi(a|s)$ as the probability that the policy $\pi$ selects action $a$ when in state $s$. Let $\Pi$ denote the set of all policies. Denote by $P_\pi$ the transition probability matrix under policy $\pi$; i.e., $P_{\pi, s, s’} = \sum_{a \in \Ac}\pi(a|s) \Pc_{s,a}(s’)$. We make the following assumption which ensures that episodes terminate.

\begin{assumption} (finite episodes) \label{assump:finite_episode}\\
For all $\pi \in \Pi$, $\lim_{h \to \infty} P_\pi^h = 0$.
\end{assumption}

Under Assumption \ref{assump:finite_episode}, the state-action value function $Q^\pi: \Sc \times \Ac \mapsto \Re$ of each policy $\pi$ is finite, where $Q^{\pi}$ is defined by 
$$Q^{\pi}(s,a) = \E_{\Mc, \pi}\left[ \sum_{h=0}^{\tau-1} r_h \Big| s_0=s, a_0 = a \right],$$ where the subscripts of the expectation indicate that state transitions and rewards are generated by MDP $\Mc$ with actions sampled by policy $\pi$.  We denote the optimal value function by $Q^*(s,a) = \max_{\pi \in \Pi} Q^\pi(s,a)$

Similar to DQN, Langevin DQN, is a value learning algorithm.  In value learning, an agent updates the parameter vector $\theta$ of a parameterized value function $Q_\theta$ as data accumulates, with an aim of estimating the optimal value function $Q^*$. At each time, an action is selected based on the current value function estimate.

A template for value learning with greedy actions is presented in Algorithm \ref{alg:episodicRL}. The algorithm begins with an empty data buffer and then iterates over episodes. In each episode, the agent updates the parameter vector based on buffered data and then applies $\epsilon$-greedy actions through termination.  The parameter update could be carried out, for example, by the DQN or Langevin DQN algorithm. These algorithms makes use of a target model with parameter vector $\theta^{\rm target}$ in their updates to maintain stability \citep{mnih2015human}. $\theta^{\rm target}$ is a snapshot of $\theta$ and updated periodically (say, every $M$ update steps).
\begin{algorithm}[!ht]
\caption{$\mathtt{greedy\_value\_rl}$}
\label{alg:episodicRL}
\begin{tabular}{lll}
\textbf{Input:} & $\theta_0$ & initial parameters \\
& $Q_\theta(s = \cdot, a = \cdot)$ & parameterized value function \\
& $\epsilon$ & exploration intensity \\
\end{tabular}
\begin{algorithmic}[1]
\STATE $\theta^{\rm target} \leftarrow \theta_0$
\STATE $\text{buffer} \leftarrow \text{Buffer.init}()$
\FOR {episode $l = 1, 2, ...$ }
\STATE $\theta,~ \theta^{\rm target} \leftarrow  \text{update}(\text{buffer}, \theta, \theta^{\rm target})$ 
\STATE $h \leftarrow 0$
\STATE observe $s^l_0$
\WHILE {$s^l_h$ is not terminal}
\STATE apply 
$$a^l_h \leftarrow 
\left\{\begin{array}{ll}
\mathtt{unif}(\arg\max_a Q_\theta(s^l_h, a)) &  \text{w.p. } 1-\epsilon \\
\mathtt{unif}(\Ac) &  \text{w.p. } \epsilon
\end{array}\right.$$
\STATE observe $r^l_h, s^l_{h+1}$
\STATE buffer.add($(s^l_h, a^l_h, r^l_h, s^l_{h+1})$)
\STATE $h \leftarrow h+1$
\ENDWHILE
\ENDFOR
\end{algorithmic}
\end{algorithm}
Observed actions, state transitions, and rewards are added to the buffer. In particular, each item in the buffer is a tuple $(s, a,r, s’)$, where $s$ is a state, $r,~s’$ are the reward and the next state observed by the agent on taking an action $a$.  The most basic version of a buffer would simply accumulate all observations.  But in practical implementations, one might design a discipline that limits the buffer size by ejecting old or randomly selected data samples.

\section{Langevin DQN} \label{sec:algorithm}

In this section we present Langevin DQN, an algorithm inspired from Langevin SGD for sequential decision problems, particularly for bandits and reinforcement learning problems. 

\subsection{Langevin DQN in Bandit Learning Context}
In this section we introduce the Langevin DQN algorithm in the bandit learning context, as it is simple and facilitates in understanding the inner workings of Langevin DQN.

Note that bandits are a special case of reinforcement learning problems, where each episode ends after a single action selection. Bandit learning algorithm follow a similar template as Algorithm \ref{alg:episodicRL}, where $\theta^{\rm target}$ is unnecessary and each data point in the buffer is of form $(s_0,a,r, s_\tau)$ with $s_0$ as a fixed start state and $s_\tau$ as a terminal state.

At the start of each episode, for a bandit problem with a Gaussian observation noise of variance $\sigma_w^2$, the Langevin DQN algorithm uses the update rule \eqref{eq:sgld} with 
$$\Lc(\theta,\tilde{\Dc}) = \sum_{(s_0,a,r,s_\tau) \in \tilde{\Dc}}\frac{(Q_\theta(s_0, a) -r)^2}{\sigma_w^2},$$
which can be replaced with an appropriate loss function for a given bandit problem.

After updating it's parameters, the Langevin DQN agent acts greedily w.r.t the current parameters without any need for additional exploration. 

\subsection{Langevin DQN in Reinforcement Learning Context}

The DQN algorithm updates the parameter vector $\theta$ by taking SGD-like steps that aim to reduce TD loss possibly summed with a regularization penalty. These SGD-like steps are similar to Algorithm \ref{alg:langDQNupdate}, but without the Gaussian noise term.  The Langevin DQN update, presented in Algorithm \ref{alg:langDQNupdate} instead uses Langevin-SGD-like steps.

\begin{algorithm}[!ht]
\caption{$\mathtt{LangevinDQN\_update}$}
\label{alg:langDQNupdate}
\begin{tabular}{lll}
\textbf{Input:} 
& $\Lc(\theta = \cdot, \theta^{\rm target} = \cdot, \Dc = \cdot)$ & TD loss \\
& $\psi(\theta = \cdot)$ & regularizer \\
& $\alpha$ & learning rate \\
& $\Dc$ & data buffer \\
& $B$ & mini-batch size \\
& $\theta^-$ & parameters \\
& $\theta^{\rm target}$ & target parameters \\
\textbf{Return:} & $\theta^+$ & new parameters
\end{tabular}
\begin{algorithmic}[1]
\STATE $\tilde{\Dc} \leftarrow \mathtt{sample\_minibatch}(\Dc, B)$
\STATE $z \leftarrow \mathtt{sample\_normal}(0, I_{\texttt{dim}(\theta^-)})$
\STATE compute increment
\begin{align}
\Delta \theta \leftarrow - \alpha \nabla_{\theta = \theta^-} & \left( \frac{1}{|\tilde{\Dc}|}\Lc(\theta, \theta^{\rm target}, \tilde\Dc) + \frac{1}{|\Dc|}\psi(\theta) \right) % \nonumber  \\ & 
+ \sqrt{\frac{2\alpha}{|\Dc|}} z \label{eq:langDQNupdate2}
\end{align}
\STATE return $\theta^- + \Delta \theta$ 
\end{algorithmic}
\end{algorithm}

For a buffer $\Dc$ with elements of form $(s,a,r, s')$, the TD loss used in algorithm \ref{alg:langDQNupdate} can be written as 
\begin{align*}
\Lc(\theta, \theta^{\rm target}, \Dc)  & = \sum_{(s,a,r, s') \in \Dc} \Lc(\theta, \theta^{\rm target}, (s,a,r,s')), \\
\text{where } ~~
\Lc(\theta, \theta^{\rm target}, (s,a,r,s')) 
& = \frac{1}{2\sigma_w^2} \left(r + \max_{a' \in \Ac}{Q}_{\theta^{\rm target}}(s', a') - Q_\theta(s,a)\right)^2,
\end{align*}
% $$\Lc(\theta, \theta^{\rm target}, \Dc)  = \sum_{(s,a,r, s') \in \Dc} \Lc(\theta, \theta^{\rm target}, (s,a,r,s')), $$
% $$\begin{array}{l}
% \text{where }
% \Lc(\theta, \theta^{\rm target}, (s,a,r,s')) % \\
% = \frac{1}{2\sigma_w^2} \left(r + \max_{a' \in \Ac}{Q}_{\theta^{\rm target}}(s', a') - Q_\theta(s,a)\right)^2,
% \end{array}$$
with an understanding that $Q_\theta(s',a') = 0$ if the transition from $s$ terminates the episode i.e., $s'$ is a terminal state.  The term
that is squared in this expression is the so-called {\it temporal difference}.

Langevin DQN is very similar to DQN, requiring very minor changes.
The primary difference lies in the fact that, while Langevin DQN updates parameters based on an expression very similar to DQN, the expression used in Langevin DQN includes an additional random perturbation term.  The only other difference is that, while $\epsilon$-greedy exploration is typically used with DQN, Langevin DQN needs no additional exploration scheme and thus it suffices to apply greedy actions. We would like to note that it is possible to design reinforcement learning algorithms that operate in continuing rather than episodic environments using an update rule similar to Langevin DQN, though we do not develop that idea here.

\subsection{Computational Experiments For Deep Exploration} \label{sec:computational_results_point_estimate}
\begin{figure}[!ht]
\centering
\includegraphics[width=0.55\columnwidth]{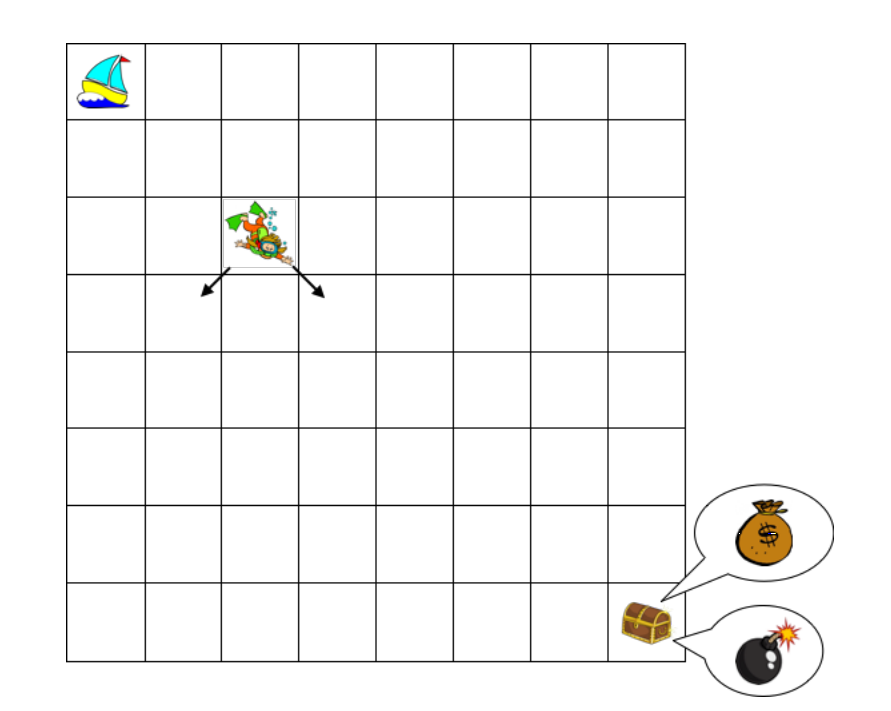}
\caption{Deep sea environment \citep{osband2017deep}}
\label{fig:deep_sea_env}
\end{figure}

To assess whether Langevin DQN achieves deep exploration, we apply it to the deep sea environment \citep{osband2017deep,osband2020behavior}. The deep sea environment with depth $N$ can be seen as an $N \times N$ two-dimensional grid in which the agent starts in the upper-right corner and should ideally reach a treasure chest at the lower-left corner as shown in Figure \ref{fig:deep_sea_env}.

At each time, the agent applies an action from $\mathcal{A} = \{0,1\}$ to move to the left or to the right cell in the next row. However, the mapping between $\{0,1\}$ to $\{\textit{left}, \textit{right}\}$ is unknown to the agent and varies across states. For simplicity, we assume that there is a treasure in the chest which is revealed to the agent when it takes the $right$ action on reaching the chest, resulting in a reward of $1$. In addition, there is a small penalty of $0.01/N$ whenever the agent takes the \textit{right} action. 

We compare the performance of Langevin DQN against traditional DQN \citep{mnih2013playing}, which uses $\epsilon$-greedy exploration. Both share Algorithm \ref{alg:episodicRL} as a template. Recall that Langevin DQN agent acts greedily according to its value function estimates which is obtained by applying iterations of Algorithm \ref{alg:langDQNupdate} in the update step in Algorithm \ref{alg:episodicRL}.

In our experiments, we use neural networks to approximate value functions. Langevin DQN and DQN each use a single neural network mapping state-action pairs to value.  We use common hyperparameters, including neural network architecture, buffer discipline, buffer size, batch size, number of updates performed between episodes, and target network update frequency, for both algorithms. This ensures that update routines impose similar computational requirements.  The exploration intensity $\epsilon$ was set to zero for Langevin DQN but tuned to optimized performance for DQN.

We used the Adam optimizer \citep{kingma2014adam} for updating parameters of DQN.  Since Langevin DQN has an additional Gaussian noise term in the update equation,  Adam could not be used in a straightforward manner.  Hence, we implemented preconditioned Langevin SGD, using the Adam optimizer as the backbone. More details about the optimizer and the exact values of hyperparameters used in our experiments are provided in Appendix \ref{app:setup}.

In order to assess and compare performance, we define a notion of learning time.  In particular, we take learning time to be the first episode at which cumulative regret is less than $0.8$ times the number of episodes.  In other words, 
$$\text{learning time} = \min\{l :\text{Regret}(l)<0.8~l\}.$$ 
The cumulative regret until the episode $l$ is defined as the difference between the maximal reward that the agent could have accumulated and the reward that the agent has accumulated over the $l$ episodes. In the case of the deep sea environment, the maximum possible reward an agent can obtain in an episode is $0.99$, and the agent observes a positive reward only when it takes the optimal actions throughout the episode and reaches the treasure chest.  In all other cases, the agent observes a non-positive reward and experiences regret of at least $0.99$.  The learning time can be thought of as the first episode by which the agent has behaved optimally for at least two tenths of past episodes, and we say that the agent has `learned' or `solved' the environment at that episode. The median learning time for an $\epsilon$-greedy strategy is lower bounded by $2^N$ \citep{osband2017deep}, an explanation for this can be found in Appendix \ref{app:dqn_langevin}. 

We carried out experiments with deep sea environments with depths varying from $10$ to $20$ in steps of $2$. We experimented with different values of $\epsilon$ for DQN and found that starting with a value of $0.5$ and decaying it optimizes the performance. For the Langevin DQN agent, we scale $\sigma_w^2$ with deep sea depths, please find the details in the Appendix \ref{app:parameters}. For each set of hyperparameters, we ran $5$ trials, each with a different random seed. In each trial, an agent operates over $5,000$ episodes. Note that as the depth of the deep sea increases, more data accumulates per episode.  Since we used a fixed mini-batch size, we scaled the number of update steps by the depth of the deep sea environment in both Langevin DQN and DQN algorithms. 

\begin{figure}[!ht]
    \centering
    \includegraphics[width=0.99\columnwidth]{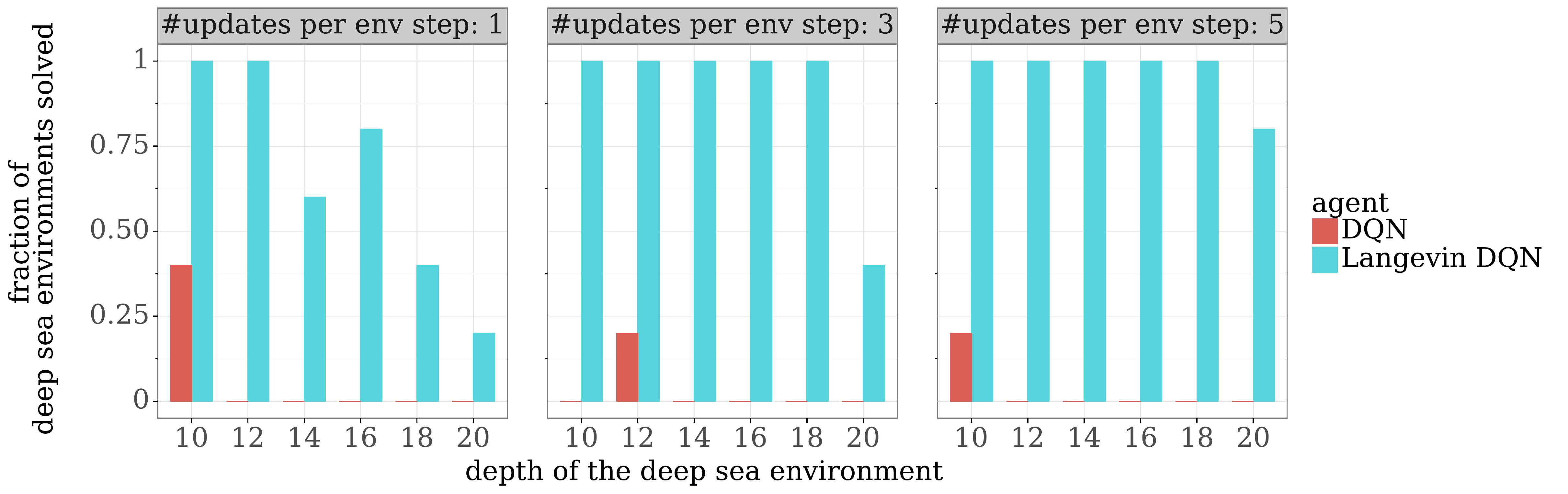}
    \caption{Performance of DQN and Langevin DQN for deep sea environments of  depths $10$ to $20$. The horizontal axis shows the depth of the deep sea environment. The vertical axis shows the fraction of seeds in which an agent has been able to solve a deep sea environment within $5000$ episodes for a specific number of update steps performed per an environment step.}
    \label{fig:dqn_vs_langevin_dqn}
\end{figure}

\begin{figure}[!ht]
    \centering
    \includegraphics[width=0.6\columnwidth]{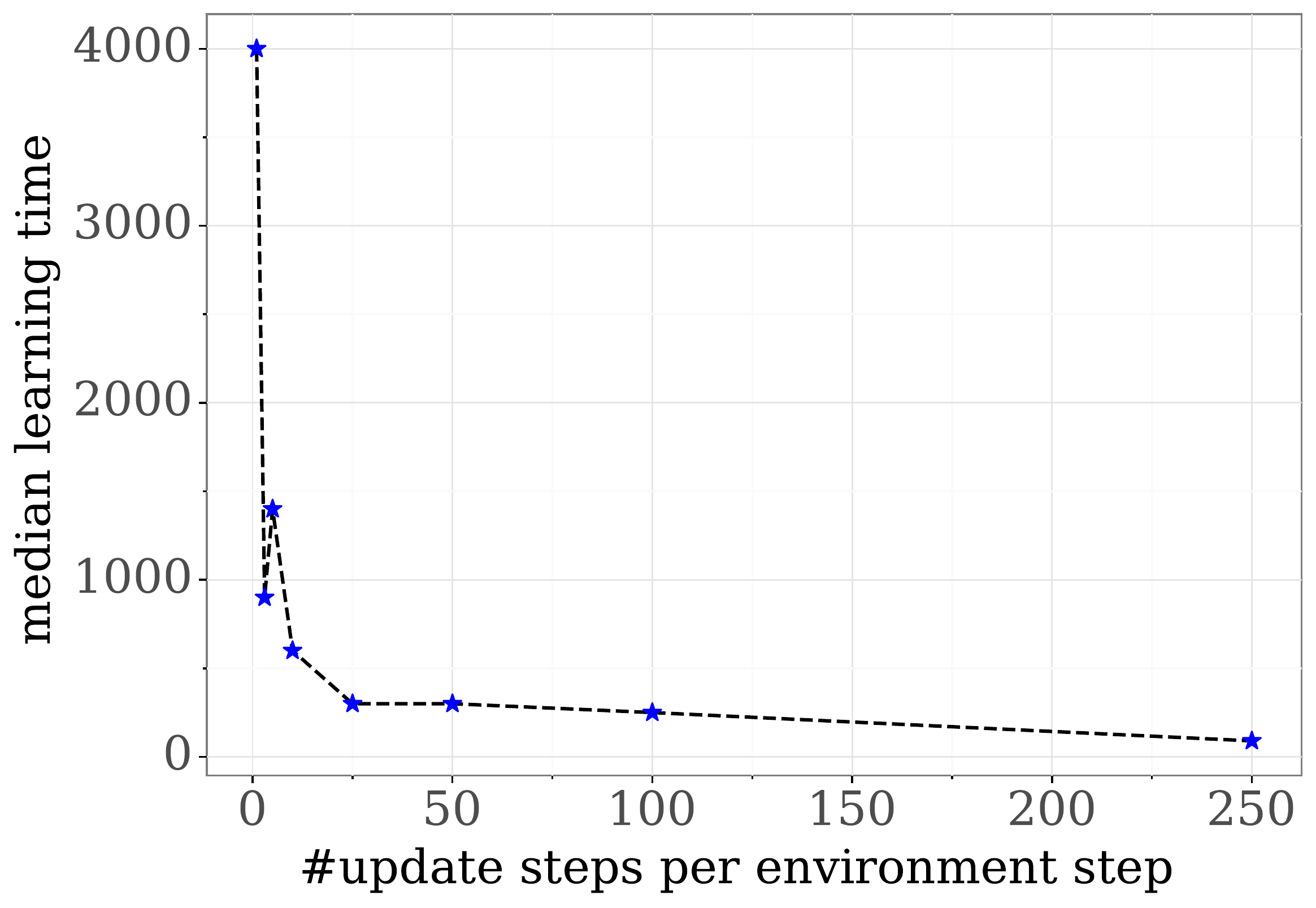}
    \caption{Median learning time of Langevin DQN on a deep sea environment of depth $10$ for varying number of update steps.}
    \label{fig:deep_sea_10-learning_time}
\end{figure}

Figure \ref{fig:dqn_vs_langevin_dqn} shows the performance of DQN and Langevin DQN algorithms on deep sea environments of different depths. The horizontal axis shows the depth of the deep sea environment. The vertical axis shows the fraction of seeds in which an agent has been able to solve a deep sea environment within $5000$ episodes for a specific number of update steps performed per an environment step. One update step per environment step is equal to $N$ number of update steps per an episode for a deep sea environment of depth $N$. Note that same hyperparameter settings are used across all the random seeds. From the figure, we can see that DQN is incapable of performing deep exploration irrespective of the number of update steps, and outperformed by Langevin DQN significantly. We can also see that as we increase the number of update steps, the performance of Langevin DQN increases. To understand this better, we plot the median learning time of Langevin DQN agent on a deep sea environment of size $10$ vs the number of update steps per environment step in Figure \ref{fig:deep_sea_10-learning_time}. From the figure, the learning time (and the number of samples required to learn) for the Langevin DQN algorithm decreases drastically with an increase in the number of update steps and goes to as low as $90$ episodes.

From these experiments we demonstrated that the Langevin DQN algorithm is capable of deep exploration, while tracking a single point estimate which is updated incrementally. We also observed, through experiments on bsuite environments \citep{osband2020behavior}, that Langevin DQN does not hurt the other capabilities of DQN. More details can be found in  Appendix \ref{app:dqn_langevin}.  

Note that the main advantage of the Langevin DQN  is its simplicity. In spite of being very similar to DQN, the Langevin DQN algorithm is capable of deep exploration. In the Section \ref{sec:ensemble_langevin}, we present Ensemble Langevin DQN which synthesizes Langevin DQN with Ensemble Sampling. This enables us to improve over the computational efficiency of Langevin DQN at the cost of using more models.

% theoretic analysis
%!TEX root = main.tex

\section{Why Langevin DQN Works}\label{sec:analysis}

In this section, we offer some insight into why Langevin DQN achieves deep exploration by perturbing parameter updates. We consider a simple example (adapted from Section 5.3.2, \citealt{osband2017deep}) to understand the role of the perturbation and how it induces deep exploration. 

To illustrate how Langevin DQN performs deep exploration, we consider a fixed horizon MDP $\mathcal{M}$ with four states $\Xc = \{1, 2, 3, 4 \}$, two actions $\Ac =\{ up, down\}$ and a horizon of $H=6$. Let $\Dc$ be the list of all transitions observed so far, and let $\Dc_{x,a} = ((\hat x,\hat a,r,x') \in \Dc : (\hat x, \hat a) = (x,a))$ contain the transitions from state-action pair $(x,a)$. Suppose $|\Dc_{4, down}|= 1$, while for every other state-action pair $(x,a) \neq (4, down)$, $|\Dc_{x,a}|$ is very large, virtually infinite. Hence, we are highly certain about the expected immediate rewards for and transition probabilities except for taking action $down$ at state $4$. Assume that this is the case for all time periods $h \in \{0,1,\dots, 5\}$.

\begin{figure}[!ht]
    \centering
    \includegraphics[width=0.5\columnwidth]{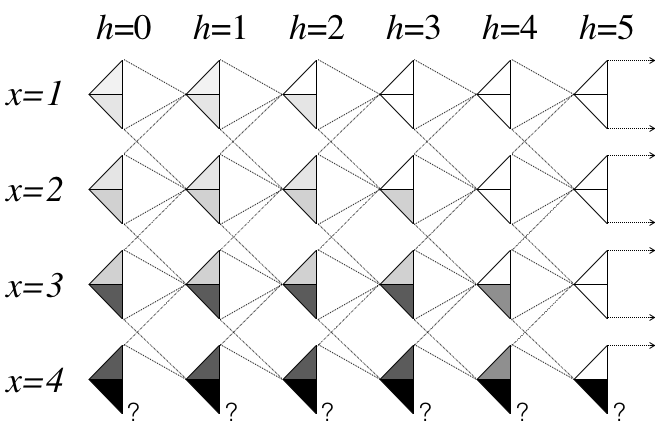}
    \caption{Example to illustrate how Langevin DQN achieves deep exploration. We can see the propagation of uncertainty from right to left in the figure. Darker shade indicates more uncertainty.}
    \label{fig:triangles}
\end{figure}

Let $Q^*_h$ denote the $Q^*$ value at time $h$. Since we are uncertain about the MDP $\Mc$, $Q^*_h$ is a random vector. Figure \ref{fig:triangles} illustrates our uncertainty in these quantities. Each triangle in row $x$ and column $h$ contains two smaller triangles that are associated with $Q^*_h$-values of $up$ and $down$ actions at state $x$. The shade on the smaller triangle shows the uncertainty in the $Q^*_h(x,a)$. Here, we consider variance as the measure of uncertainty. The dotted lines show plausible transitions, except at $(4, down)$. Since we are uncertain about $(4, down)$, any transition is plausible. 

At $h=H-1=5$, since $Q^*_5$ is only affected by the immediate rewards, the only uncertain quantity is $Q^*_5(4, down)$. If we move one step back to $h=4$, we are  highly uncertain about $Q_4^*(4, down)$ and slightly uncertain about $Q^*_4(3, down)$ and $Q_4^*(4, up)$. This is because of the propagation of uncertainty from $(4, down)$ at $h=5$ to $(3, down)$ and $(4, up)$ at $h=4$. The uncertainty propagates backward in time, which can be visualized as progressing leftward in Figure \ref{fig:triangles}. Specifically, we can see that the uncertainty in $Q^*_5(4, down)$ can even influence the uncertainty in $Q^*_0(1, up)$.

Now, let us look at $\hat{Q}_h$, $Q$-values generated by Langevin DQN using update equation \eqref{eq:langDQNupdate2} with $\alpha = \epsilon |\Dc|$. Here, we provide an argument for why the variance of $\hat{Q}_h(x,a)$ grows and shrinks with the variance of $Q^*_h(x,a)$ for all state-action pairs $(x,a)$. For simplicity, let the entire buffer be used instead of a mini-batch, $\psi$ be $0$, target $Q$-values are set as the current $Q$-value estimates, and a tabular representation be used for the value functions. In particular, parameter vector is the concatenation of $\hat{Q}_0, \hat{Q}_1, \dots \hat{Q}_5 \in \Re^{|\Xc|\times |\Ac|}$. From these assumptions, the update equation \eqref{eq:langDQNupdate2} at the $k$th update is given by 
\begin{align}
    \hat{Q}^k_h(x,a) = &  \hat{Q}^{k-1}_h(x,a) % \nonumber \\    & 
    -\epsilon \frac{1}{\sigma_w^2}\sum_{(x,  a, r, x') \in \Dc_{x,a}}\Big({\hat{Q}^{k-1}_h}(x, a) - r % \nonumber \\     & ~~~~~~~~ 
    - \max_{a' \in \Ac} \hat{Q}^{k-1}_{h+1}(x',a')\Big) + \sqrt{2\epsilon}z^k_h \label{eq:tabular_lang}
\end{align}
with $\hat{Q}^k_6(x,a) = 0~\forall x \in \Xc, a \in \Ac, k \geq 0$.

When $|\Dc_{x,a}|$ is large, the perturbation term is dominated by the loss term and $\hat{Q}^k_h(x,a)$ remaining almost unchanged and close to $\mathbb{E}[r_{t+1} + \max_{a' \in \Ac} \hat{Q}^{k-1}_{h+1}(x_{h+1},a')|x_h=x,a_h=a]$. This is true for all state-action pairs except $(4, down)$. Since $|\Dc_{4, down}|=1$, the value of $\hat{Q}^k_h(4,down)$ will exhibit high variation.

Because of this, at $h=5$, $\hat{Q}_5^k(x,a)$ is highly concentrated around $Q^*_5(x,a)$ for all $(x,a) \neq (4, down)$, and $\hat{Q}_5^k(4, down)$ will have high variance. At $h=4$, $\hat{Q}_4^k(4, up)$ and $\hat{Q}_4^k(3, down)$ also start exhibiting variance after a few updates because of the propagation of uncertainty from $\hat{Q}_5^k(4, down)$ through \eqref{eq:tabular_lang}. Working backward in time, leftward in Figure \ref{fig:triangles}, we can see that the uncertainty propagates backward, as shown by the shaded areas in the triangles.

% The update equation \eqref{eq:tabular_lang}, can be seen as a discretized version of a SDE
% \begin{align*} d\hat{Q}^t_h(x,a) = &- \frac{1}{\sigma_w^2}\sum_{(\tilde x, \tilde a, r, x') \in \Dc_{x,a}}\Big(\hat{Q}^t_h(\tilde x, \tilde a) - r  \nonumber \\
%     & ~~~~~  - \max_{a' \in \Ac} \hat{Q}^t_{h+1}(x',a')\Big)dt  + \sqrt{2} dB_h^t,
% \end{align*}
% where $B_h^t$ is a standard Brownian motion in $\Re^d$. This SDE can also help us understand the uncertainty propagation. The uncertainty in $\hat{Q}_5(4, down)$ contributes to the uncertainty in $\hat{Q}_4(3, down)$ and $\hat{Q}_4(4, up)$.  As we go backward in time, the uncertainty propagates as shown by the shaded triangles in Figure \ref{fig:triangles}.  

This uncertainty drives exploration. A high variance in $\hat{Q}_h(x,a)$  generates optimistic $\hat{Q}_h(x,a)$ in some episodes,  incentivizing the agent to try those actions. Note that the uncertainty in the state-action value exists not only because of the uncertainty in the immediate reward and transitions but also from the propagation of the uncertainty from multiple time periods. This process leads to deep exploration. 

In order to validate this insight, let us look at the uncertainty at different states of deep sea environment (Section \ref{sec:algorithm}) as the Langevin DQN agent interacts over multiple episodes.

\begin{figure}[!ht]
    \centering
    \includegraphics[width=0.9\columnwidth]{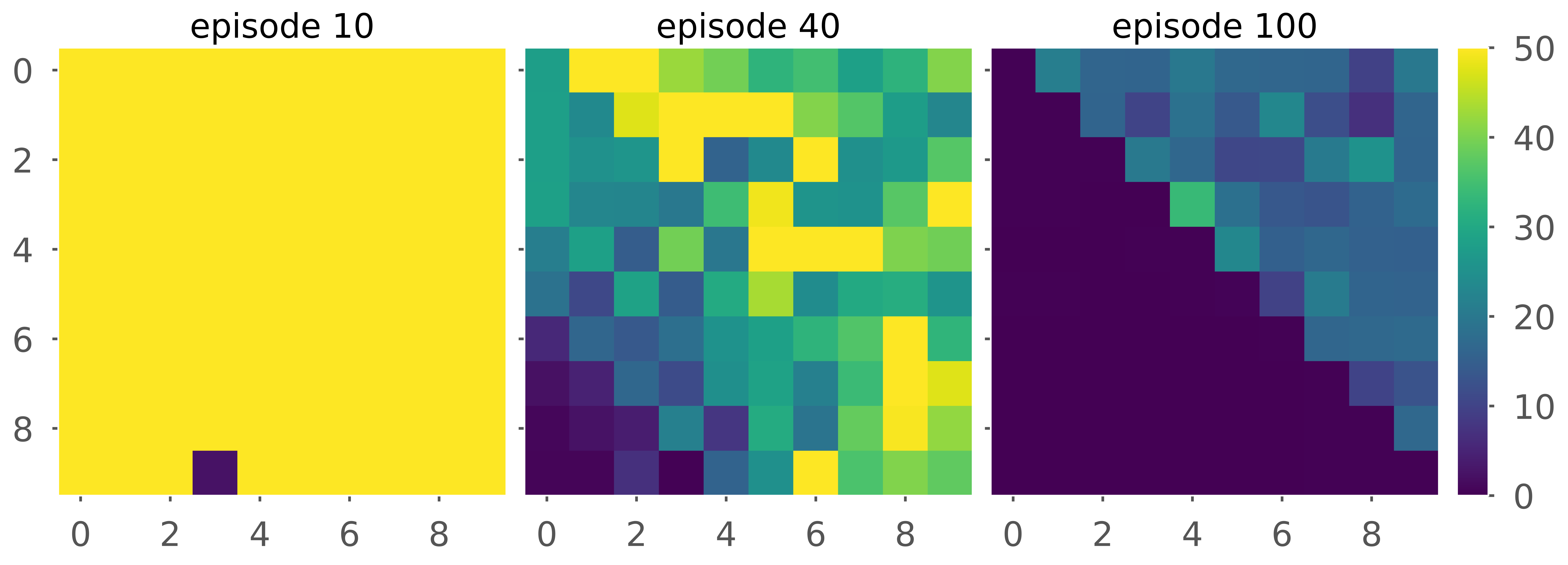}
    \caption{Uncertainty of Langevin DQN at different states of a deep sea environment of depth $10$ when run for $250$ update steps per environment step. 
    We define uncertainty of a state as the sum of standard deviations of Q-values across all actions at that state.
    }
    \label{fig:std_dev_deep_sea}
\end{figure}

Figure \ref{fig:std_dev_deep_sea} shows the uncertainty of Q-values of Langevin DQN at different states of the deep sea environment of depth $10$. The parameters used for Langevin DQN are same as the ones used in Section \ref{sec:computational_results_point_estimate} and can be found in Appendix \ref{app:dqn_langevin}. At the start of each episode, the agent performs $250$ update steps per environment step, i.e., $2500$ update steps. The uncertainty at each state is calculated by summing over the standard deviation of Q-values across all actions at that state, where the standard deviation of a Q-values at a state-action pair is calculated empirically based on the Q-values of the past $10$ episodes. We have clipped the standard deviation values at $50$ for better visualization. Note that a state is uncertain if the Q-values of any action at that state are uncertain, and in a deep sea environment, the agent cannot reach the states in the upper right triangle of the grid.
From the figure, we can see that at episode $10$, the agent is uncertain about almost the entire state space, except for a single state. By episode $40$, the agent has observed few states at the bottom left corner, which reduces uncertainty in Q-values at those states; however, since some trajectories from the top left states lead to bottom right states which are unexplored, the Q-values at the top left states are still uncertain even though they have been visited several times. This is consistent with intuition provided in this section. As the agent interacts with the environment, it explores, resolves its uncertainty, and finally learns an optimal policy.

% Ensemble Langevin DQN 
%!TEX root = main.tex

\section{Ensemble Langevin DQN}\label{sec:ensemble_langevin}

In order to increase the computational efficiency, we combined Langevin DQN with Ensemble sampling \citep{lu2017ensemble} to produce Ensemble Langevin DQN. This also enabled us to use prior networks \citep{osband2018randomized}. Ensemble Langevin DQN  uses an ensemble of models and updates each model independently using the Langevin DQN update rule in Algorithm \ref{alg:langDQNupdate}. Models in the ensemble differ by initialization, prior network \citep{osband2018randomized} and the Gaussian noise term in the gradients. 

Ensemble Langevin DQN and Boot DQN (an ensemble sampling algorithm, \citet{osband2017deep}) are very similar, and a common template is provided in Algorithm \ref{alg:ensemble}. Both the algorithms maintain an ensemble of models. At the start of each episode, an agent updates the ensemble, picks a model uniformly at random, and acts greedily based on the picked model and appends the data in that episode into the buffer.
\begin{algorithm}[!ht]
\caption{$\mathtt{ensemble\_value\_rl}$}
\label{alg:ensemble}
\begin{tabular}{lll}
\textbf{Input:} & $\{\theta^0_1, \ldots, \theta^0_M\}$ & initial parameters \\
& $Q_\theta(s = \cdot, a = \cdot)$ & parameterized value function \\
& $p$ & sample insertion probability \\
\end{tabular}
\begin{algorithmic}[1]
\FOR{$k  = 1,\ldots, M$}
\STATE $\theta^{\rm target}_k \leftarrow \theta^0_k$
\STATE $\text{buffer}_k \leftarrow \text{Buffer.init}()$
\ENDFOR
\FOR {episode $l = 1, 2, ...$ }
\FOR{$k  = 1,\ldots, M$}
\STATE $\theta_k,~ \theta^{\rm target}_k \leftarrow  \text{update}(\text{buffer}_k, \theta_k, \theta^{\rm target}_k)$ 
\ENDFOR
\STATE $m \sim {\rm Unif}(\{1,\ldots,M\})$
\STATE $h \leftarrow 0$
\STATE observe $s^l_0$
\WHILE {$s^l_h$ is not terminal}
\STATE apply 
$a^l_h \leftarrow \mathtt{unif}(\arg\max_a Q_{\theta_m}(s^l_h, a))$
\STATE observe $r^l_h, s^l_{h+1}$
\FOR{$k  = 1,\ldots, M$}
\IF{${\rm Bernoulli}(p)$ is $1$}
\STATE $\text{buffer}_k.\text{add}((s^l_h, a^l_h, r^l_h, s^l_{h+1})$)
\ENDIF
\ENDFOR
\STATE $h \leftarrow h+1$
\ENDWHILE
\ENDFOR
\end{algorithmic}
\end{algorithm}
Key differences between Ensemble Langevin DQN and Boot DQN lie in how the models are updated and how the new data samples are added into the buffer. Each model in Ensemble Langevin DQN is updated using Algorithm \ref{alg:langDQNupdate}, while Boot DQN is updated using a similar algorithm without the Gaussian noise term. In addition, Boot DQN uses bootstrapping techniques to perturb the data, which involves maintaining a different buffer for each model of the ensemble. Once a data point is observed, it is added into a buffer with probability $p$. In case of Ensemble Langevin DQN, $p=1$, i.e., all models use the same buffer, which contains all data.

Through the computational results presented in Section \ref{sec:computational_results} and Appendix \ref{app:boot_langevin}, we observed that, in spite of being very similar to Boot DQN, Ensemble Langevin DQN offers greater ability to perform deep exploration.  

\subsection{Computational results} \label{sec:computational_results}

In order to access the ability of Ensemble Langevin DQN to perform deep exploration, we apply it to deep sea environment described in Section \ref{sec:computational_results_point_estimate}. 

\begin{figure}[!h]
    \centering
    \includegraphics[width=0.8\columnwidth]{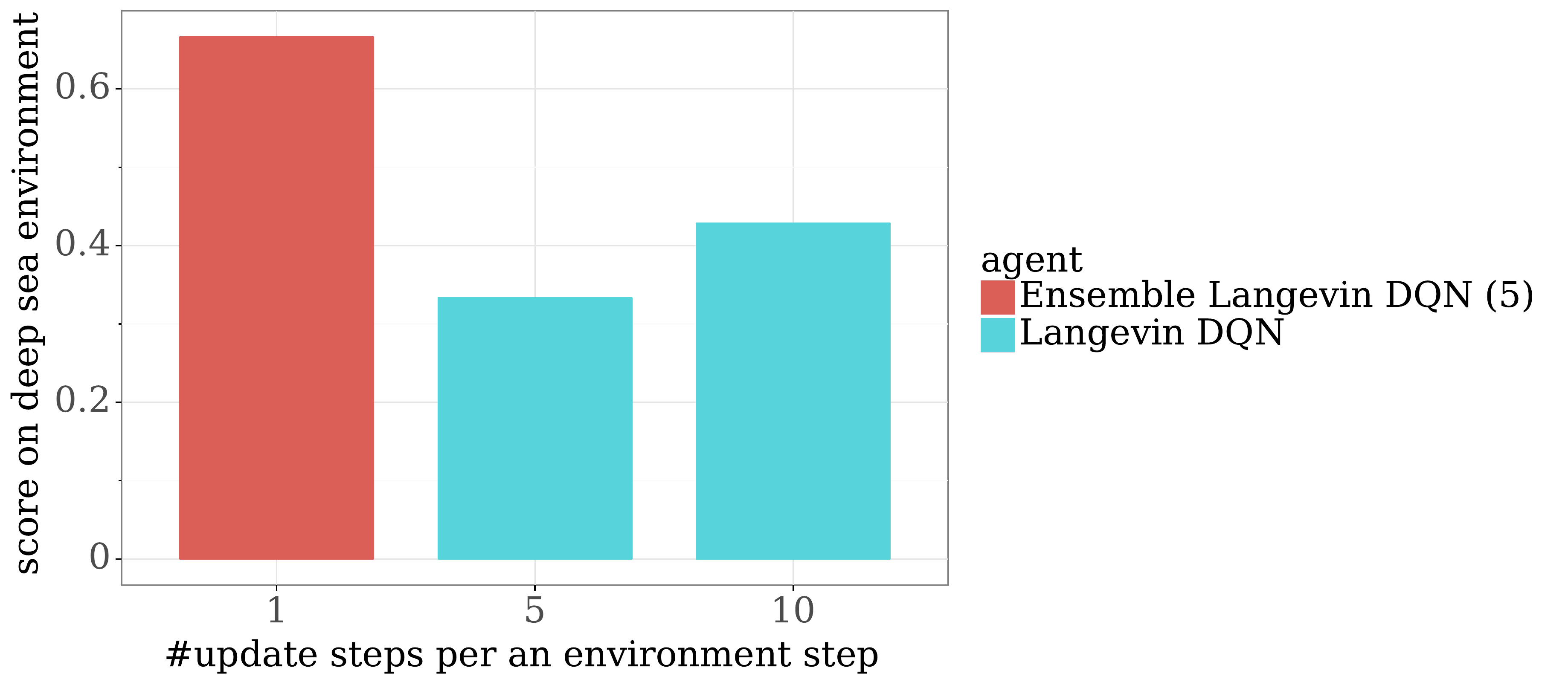}
    \caption{Performance of Langevin DQN and Ensemble Langevin DQN (with $5$ models) for different number of update steps. Note that computation required for an ensemble of size $5$ is equivalent to that of a single model with $5$ times the number of update steps.}
    \label{fig:langevin_vs_ensemble_langevin}
\end{figure}

Figure \ref{fig:langevin_vs_ensemble_langevin} shows the performance of Langevin DQN and Ensemble Langevin DQN over the deep sea environment. The vertical axis shows the score of an agent on the deep sea environments in bsuite. We define this score as the fraction of deep sea environments, with depths $10$ to $50$ in increments of $2$, that the agent solved in less than $5000$ episodes.  The horizontal axis shows the number of update steps per an environment step. Each update step of Ensemble Langevin DQN ($5$), which uses $5$ models, uses same computation as $5$ update steps of Langevin DQN. From the figure, we can see that Ensemble Langevin DQN improves over the ability of Langevin DQN to perform deep exploration for the same amount of compute at the cost of using more models.

Next we compare Ensemble Langevin DQN with Boot DQN and show that Langevin DQN can outperform Boot DQN. We use same hyperparameter settings such as neural network architecture, buffer discipline, buffer size, batch size, target update frequency to ensure that both Boot DQN and Ensemble Langevin DQN uses similar computation per model. The value of sample insertion probability $p$ is set to $0.5$ for Boot DQN and $1$ for Ensemble Langevin DQN. More details about the hyperparameter settings can be found in Appendix \ref{app:param_boot_dqn_langevin}.

\begin{figure}[!ht]
    \centering
    \includegraphics[width=0.8\columnwidth]{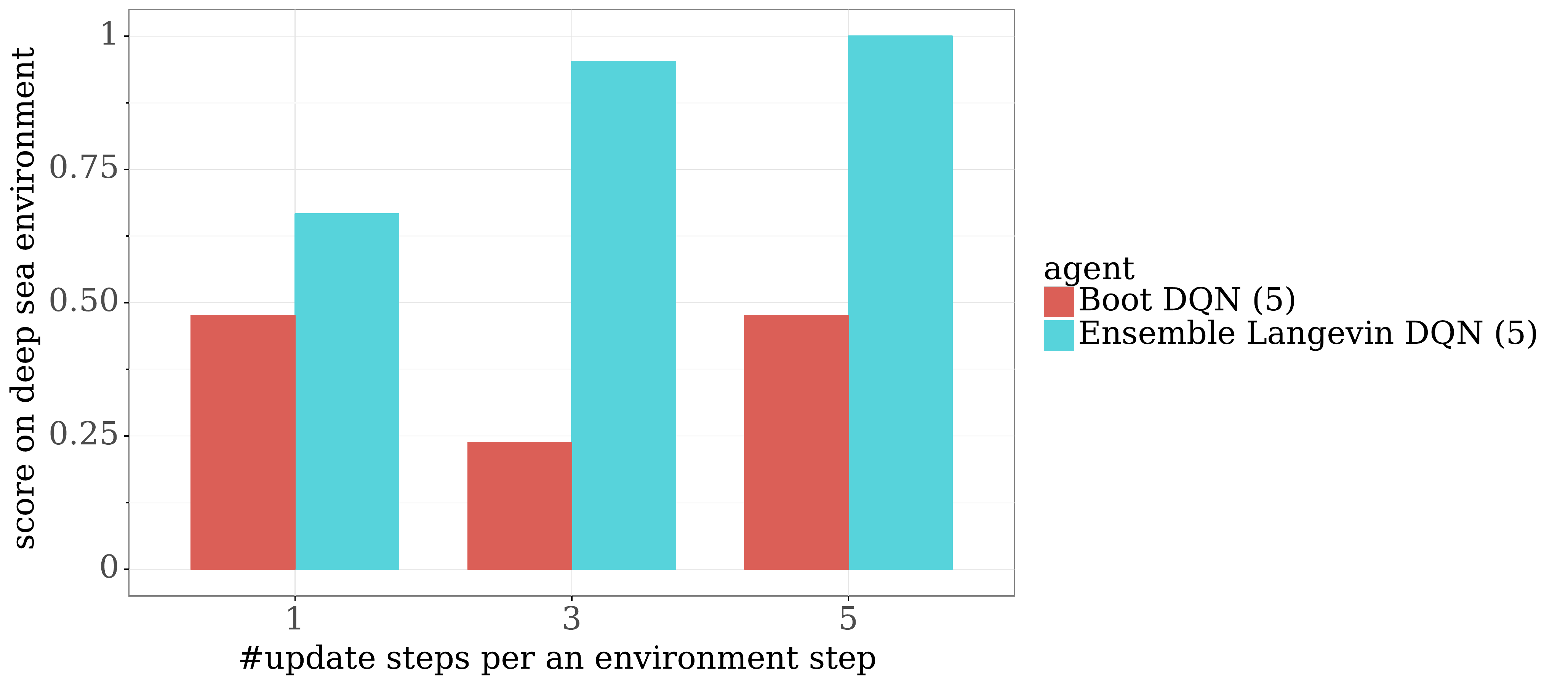}
    \caption{Comparing performance of Ensemble Langevin DQN and Boot DQN across a different number of update iterations. Both the agents use an ensemble of size $5$. Score on the vertical axis is the fraction of deep sea environments in bsuite that the agent has solved.}
    \label{fig:deep_sea_num_sgd}
\end{figure}

We have compared the performance of Boot DQN and Ensemble Langevin DQN, both with an ensemble of size $5$, across a different number of update iterations in Figure \ref{fig:deep_sea_num_sgd}. 
The horizontal axis shows the number of update steps per an environment step and vertical axis shows the score on the deep sea environment, which is the fraction of deep sea environments in the Bsuite that the agent has solved within $10,000$ episodes. Bsuite has $21$ deep sea environments with depth varying from $10$ to $51$ in steps of $2$. Note that the difficulty to solve a deep sea environment increases exponentially with an increase in the depth of the deep sea.  Observe that the performance of Boot DQN does not improve even after increasing the number of update iterations.

\begin{figure}[!ht]
    \centering
    \includegraphics[width=0.8\columnwidth]{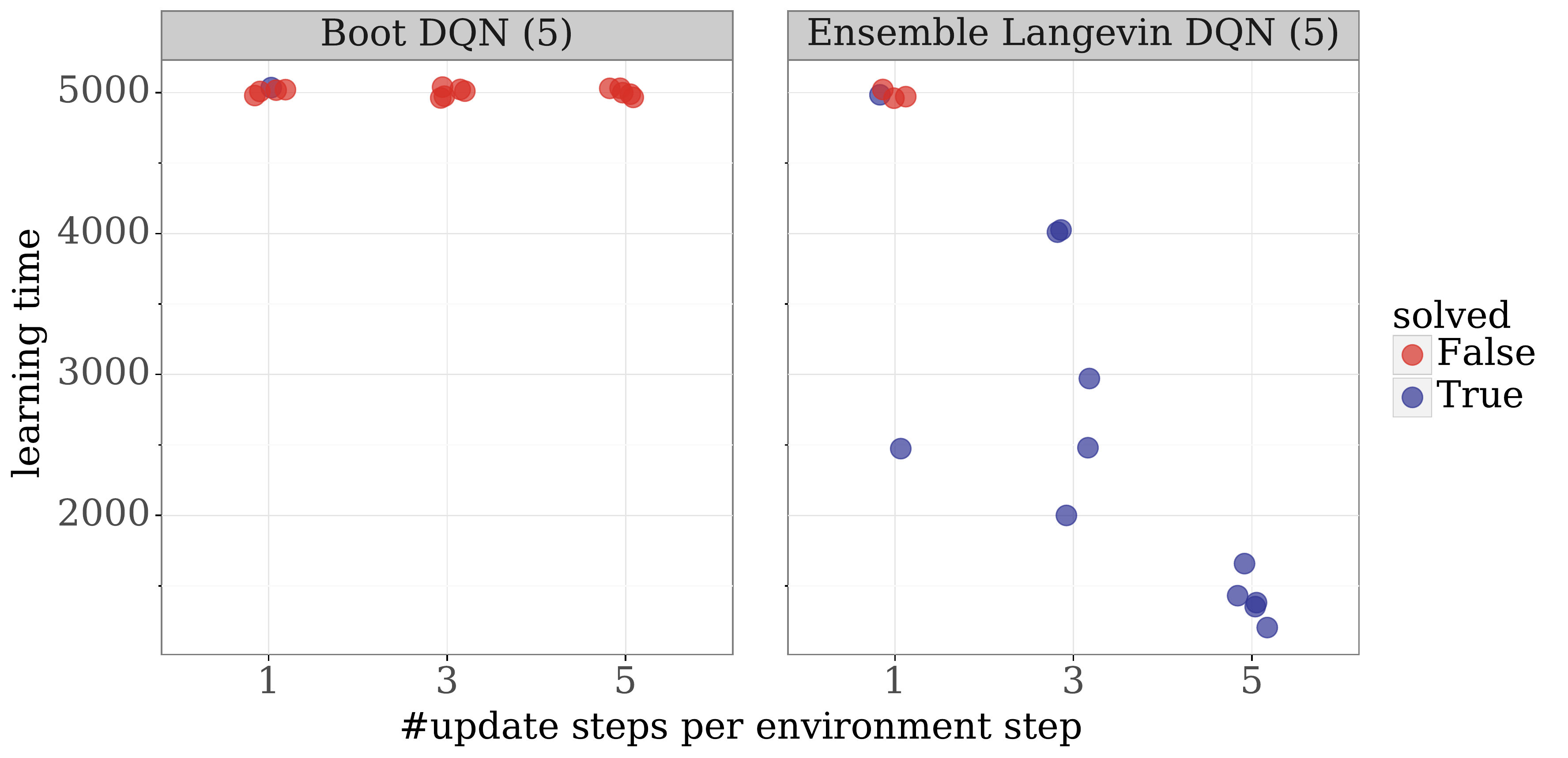}
    \caption{Comparing learning time of Ensemble Langevin DQN and Boot DQN on deep sea environment of depth $50$. The horizontal axis shows the number of update steps per environment step, and the vertical axis shows the learning time of an agent.}
    \label{fig:deep_sea_50_num_sgd_learning_times}
\end{figure}

Figure \ref{fig:deep_sea_50_num_sgd_learning_times} compares the learning times of Ensemble Langevin DQN with Boot DQN, both with an ensemble of size $5$, when applied to a deep sea environment of depth $50$ across a different number of update iterations. 
The horizontal axis shows the number of update steps per environment step, and the vertical axis shows the learning time of an agent as defined in Section \ref{sec:computational_results_point_estimate}. For each value of update iterations, we ran $5$ trails, each with a different random seed, each operated over $5000$ episodes. For comparison, the median number of episodes required by a DQN agent to learn a deep sea environment of depth $50$, as derived in Section \ref{sec:computational_results_point_estimate}, is given by  $2^{50}=1.25 \times 10^{15}$.  From the figure, it is clear that Ensemble Langevin DQN outperforms Boot DQN significantly in terms of its ability to perform deep exploration, and Ensemble Langevin DQN can learn an optimal policy sooner on increasing the number of update iterations unlike Boot DQN. We can also see that the performance is robust across different seeds.

Please refer to Appendix \ref{app:boot_langevin} for a more comprehensive set of experiments, on a wide range of environments in Bsuite \citep{osband2020behavior}, to test the overall performance of the Ensemble Langevin DQN. 

\section{Conclusion} \label{sec:conclusion}

In this paper, we introduced Langevin DQN, a slight variation of DQN that achieves deep exploration using a single point estimate of the value function. We showed this through computational experiments and provided insights into how Langevin DQN accomplishes this. We also introduced Ensemble Langevin DQN to improve computation efficiency over Langevin DQN and perform deep exploration at the cost of using more models. The key point we want to convey is that Langevin DQN is capable of deep exploration with a single point estimate with no additional complexity to account for epistemic uncertainty.

\section*{Acknowledgments}

We thank Vivek Borkar, Xiuyuan Lu, Morteza Ibrahimi and Ian Osband for helpful discussions and pointing out errors and ambiguities in earlier drafts. This research was supported by the Army Research Office (ARO) under award W911NF2010055.

\bibliography{references}

% \newpage
%!TEX root = main.tex

\appendix
\section{Experimental Setup} \label{app:setup}

Code for Langevin DQN can be found at \url{https://github.com/vik0/LangevinDQN}. We implement most of our code in TensorFlow using the source code of Bsuite \citep{osband2020behavior} at \url{https://github.com/deepmind/bsuite}. Bsuite is a collection of carefully designed experiments used to investigate the core capabilities of reinforcement learning agents. For more details, please refer to \citet{osband2020behavior}. 

In Appendix \ref{app:setup}, we provide pseudo-code for the optimizer used by Langevin DQN and the parameters used in different experiments. In Appendix \ref{app:additional_results}, we present results which complement the computational results in Sections \ref{sec:computational_results_point_estimate} and \ref{sec:computational_results}, further strengthening our claims on the performance of Langevin DQN.

\subsection{Optimizer Details} \label{app:optimizer}
In our experiments, we use neural networks to represent the value functions. It is well known that simple optimizers like SGD do not work well with neural networks because of the complicated loss manifolds generated by them. Because of this, we use Adam optimizer \citep{kingma2014adam} for DQN and Boot DQN. However, we cannot use Adam optimizer for Langevin DQN as the update rule has an additional Gaussian noise term along with the stochastic gradient term. Hence, we use a preconditioned version of Langevin SGD \eqref{eq:langDQNupdate2}, using Adam optimizer as the backbone, for both Langevin DQN and Ensemble Langevin DQN. The pseudo-code for updating a parameter $\theta$ using the Langevin SGD optimizer is provided in Algorithm \ref{alg:langevin_sgd_optimizer}. $\hat{g}^2$ used in the algorithm is the element wise square of the stochastic gradient $\hat{g}$.
\begin{algorithm}[!ht]
\caption{$\mathtt{Langevin\_SGD\_optimizer}$}
\label{alg:langevin_sgd_optimizer}
\begin{tabular}{lll}
\textbf{Input:} & $\alpha$ & learning rate \\
& $\beta_1, \beta_2$ & exponential decay rate of $1$st and $2$nd moments of gradients \\
& $\epsilon$ & small constant for numerical stability \\
& $\hat{g}$ & stochastic gradient of the loss \\
& $k$ & iteration number \\
& $\theta_{k-1}$ & parameters at $k-1^{th}$ iteration \\
& $m_{k-1}, v_{k-1}$ & first and second raw moment estimates at $k-1^{th}$ iteration \\
\textbf{Return:} & $\theta_k$ & new parameters \\
& $m_k, v_k$ & new first and second raw moment estimates
\end{tabular}
\begin{algorithmic}[1]
\STATE Sample a Gaussian random vector with same dimension as $\theta$: $z \sim \mathcal{N}(0, I)$
\STATE Update: \begin{align*}
    \alpha_k & = \alpha \left( \frac{\sqrt{1 - \beta_2^k}}{1 - \beta_1^k} \right) \\    
    m_k & = \beta_1 m_{k-1} + (1 - \beta_1) \hat{g} \\
    v_k & = \beta_2 v_{k-1} + (1 - \beta_2)  \hat{g}^2 \\
    \theta_k & = \theta_{k-1} - \alpha_k  \frac{m_k}{\sqrt{v_k} + \epsilon} +\sqrt{\frac{2 \alpha_k}{\sqrt{v_k} + \epsilon}} z
\end{align*}
\STATE return $\theta_k, m_k, \text{ and } v_k$
\end{algorithmic}
\end{algorithm}

The update rule used by Langevin DQN and Ensemble Langevin DQN in our experiments is 
$$ \Delta \theta \leftarrow - \alpha \nabla_{\theta = \theta^-}  \left( \sum_{(s,a,r, s') \in \Dc} \frac{\left(r + \max_{a' \in \Ac}{Q}_{\theta^{\rm target}}(s', a') - Q_\theta(s,a)\right)^2 }{| \tilde{\Dc}|} + \frac{\sigma_w^2}{|\Dc|}\psi(\theta) \right)  + \sqrt{\frac{2\alpha\sigma_w^2}{|\Dc|}} z. 
$$ 

Note that this update rule is obtained using a TD loss in the update rule \eqref{eq:langDQNupdate2} and then scaling the learning rate with $\sigma_w^2$. This enabled us to use a single learning rate for different values of $\sigma_w^2$ in our experiments. It is worth noting that we tried both intra-episodic and inter-episodic updates and did not observe a significant difference in the performance. We present the results from inter-episodic updates in this paper.

 We use $\psi(\theta) = \lambda \|\theta\|^2$ in our experiments.

\subsection{Parameters Used} \label{app:parameters}

In all our experiments, we used $2$ layer neural networks to represent value functions. We use same hyperparameter values such as number of hidden units, buffer size, buffer discipline, batch size, frequency of updating the target networks in order to ensure that different agents use similar compute per update iteration per model. We tuned over the hyperparameters which might have a significant impact on the performance of an agent, and the rest of hyperparameters were set to the default values from the Bsuite code base. Here are the parameters we used for different experiments. 

\hspace{4mm}
\subsubsection{DQN vs Langevin DQN} \label{app:param_dqn_langevin}
The parameters for DQN and Langevin DQN are provided in Table \ref{table:dqn_langevin}.
\begin{table}[!h]
    \centering
    \begin{tabular}{| p{0.4\columnwidth} | p{0.20\columnwidth} | p{0.2\columnwidth} |}
    \hline 
      parameter  &  DQN & Langevin DQN \\
    \hline
    number of hidden layers & $2$ & $2$ \\
    number of units per hidden layer & $50$ & $50$ \\
    activation function & ${\rm ReLU}$ & ${\rm LeakyReLU}$ \\
     learning rate & $0.001$ & $0.01$ \\
     mini-batch size & $128$ & $128$ \\
     buffer size (in samples) & $100,000$ & $100, 000$ \\
     target update frequency (in number of update steps) & $4$ & $4$ \\
     $\sigma_w^2$ & $0.0$ & $0.005$ \\
     $\lambda$ for $\psi(\theta) = \lambda\|\theta\|^2$ & $0$ & $1$ \\
     $\epsilon$ & $0.05$ & $0$ \\
    \hline
    \end{tabular}
    \caption{Parameters used in experiments comparing DQN and Langevin DQN.}
    \label{table:dqn_langevin}
\end{table}

These are the default parameter values used for the experiments in Section \ref{sec:computational_results_point_estimate} and Appendix \ref{app:dqn_langevin}. In these experiments, DQN uses a ReLU activation function and Langevin DQN uses a LeakyReLU activation function, with a negative slope coefficient of $0.1$. We found that Langevin DQN works with smaller negative slope coefficients, but requires a higher number of update iterations. 

For generating Figure \ref{fig:dqn_vs_langevin_dqn}, we used $\sigma_w^2 = 0.005 \times {\rm depth ~ of ~ the ~ deep ~ sea ~ environment}$ for Langevin DQN, and for DQN we tried different values of $\epsilon$ and found that starting with $\epsilon=0.5$ and decaying it with time gave the best performance. Even with a fixed value of $\sigma_w^2=0.005$, the performance of Langevin DQN was only slightly lower than using a scaled $\sigma_w^2$. We also tried using a larger learning rate and LeakyReLU activation for DQN, but it did not improve the performance of DQN.

For computational experiments on all Bsuite environments, presented in Appendix \ref{app:dqn_langevin}, we used a fixed value of $\epsilon=0.05$ for DQN, which was the default value used in Bsuite code for DQN, and we used a fixed $\sigma_w^2=0.005$ for Langevin DQN. We used a fixed set of hyperparameters to generate results on all Bsuite environments, i.e., we did not change the hyperparameters for different environments. This tests for the robustness of the algorithm at the specified parameters across different Bsuite environments.

\subsubsection{Boot DQN vs Ensemble Langevin DQN} \label{app:param_boot_dqn_langevin}

For each model of Boot DQN and Ensemble Langevin DQN, we used the parameters provided in Table \ref{table:boot_langevin}.
\begin{table}[!ht]
    \centering
    \begin{tabular}{| p{0.4\columnwidth} | p{0.20\columnwidth} | p{0.2\columnwidth} |}
    \hline 
      parameter  &  Boot DQN & Ensemble Langevin DQN \\
    \hline
    number of hidden layers & $2$ & $2$ \\
    number of units per hidden layer & $50$ & $50$ \\
    activation function & ${\rm ReLU}$ & ${\rm ReLU}$ \\
     learning rate & $0.001$ & $0.001$ \\
     mini-batch size & $128$ & $128$ \\
     buffer size (in samples) & $100,000$ & $100, 000$ \\
     target update frequency (in number of update steps) & $4$ & $4$ \\
     $\sigma_w^2$ & $0.0$ & $0.0001$ \\    
     $\lambda$ for $\psi(\theta) = \lambda\|\theta\|^2$ & $0$ & $0$ \\
     prior scale & $3$ & $3$ \\
     Insertion probability & $0.5$ & $1$ \\
    \hline
    \end{tabular}
    \caption{Parameters used in experiments comparing Boot DQN and Ensemble Langevin DQN.}
    \label{table:boot_langevin}
\end{table}

Each model of the Ensemble Langevin DQN is identical to a model of Boot DQN except for the Gaussian noise term in the update step and sample insertion probability. We use prior networks \citep{osband2018randomized} for both Boot DQN and Ensemble Langevin DQN. We used these parameters for experiments in Section \ref{sec:computational_results} and Appendix \ref{app:boot_langevin}.

\subsubsection{Langevin DQN vs Ensemble Langevin DQN}

For Figure \ref{fig:langevin_vs_ensemble_langevin}, we used the parameters specified in Appendix \ref{app:param_dqn_langevin} for Langevin DQN and the parameters specified in Appendix \ref{app:param_boot_dqn_langevin} for Ensemble Langevin DQN.

% ================================================================================

\section{Additional Results} \label{app:additional_results}

\subsection{DQN vs Langevin DQN} \label{app:dqn_langevin}

\subsubsection{Lower Bound on Median Learning Time of DQN}
It is easy to establish a lower bound on the median learning time of dithering exploration schemes like $\epsilon$-greedy on deep sea environment. When the agent starts with an uninformative value function, the agent needs to reach the treasure chest to learn the optimal value function. Let us ignore the penalty of $-0.01/N$ on taking the ${\it right}$ action. Let $l^*$ be the first episode at which the agent reaches the treasure chest. Under the $\epsilon$-greedy strategy, for any episode before $l^*$, all actions are sampled uniformly at random irrespective of the exploration intensity $\epsilon$. Thus, the probability of reaching the treasure is $1/2^N$. Hence, $l^*$ is a geometric random variable with the probability of success as $1/2^N$. If we account for the penalty for taking the $\textit{right}$ action, the probability of success becomes even lower. Thus, the median learning time for an $\epsilon$-greedy strategy is lower bounded by $2^N$. 

\subsubsection{Experiments on Bsuite}

In addition to the experiments on deep sea environment, presented in Section \ref{sec:computational_results_point_estimate}, we have also conducted experiments on all the bsuite environments. In these experiments, we used a fixed hyperparmeter settings for all the bsuite environments. The hyperparameter used for these experiments can be found in the Appendix \ref{app:param_dqn_langevin}. 

\begin{figure}[!ht]
    \centering
    \includegraphics[width=0.9\columnwidth]{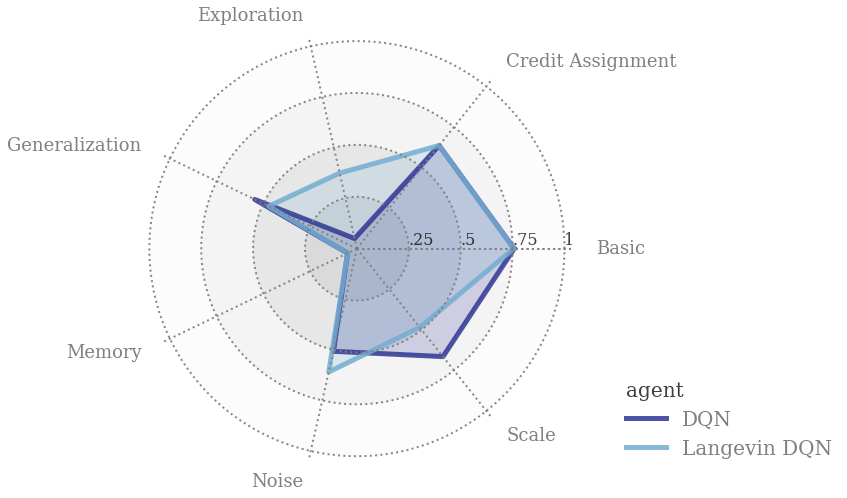}
    \caption{Performance of Langevin DQN and DQN on different categories of Bsuite environments. Both the agents use $5$ update steps per environment step.}
    \label{fig:radar-dqn_langevin}
\end{figure}

\begin{figure}[!ht]
    \centering
    \includegraphics[width=0.9\columnwidth]{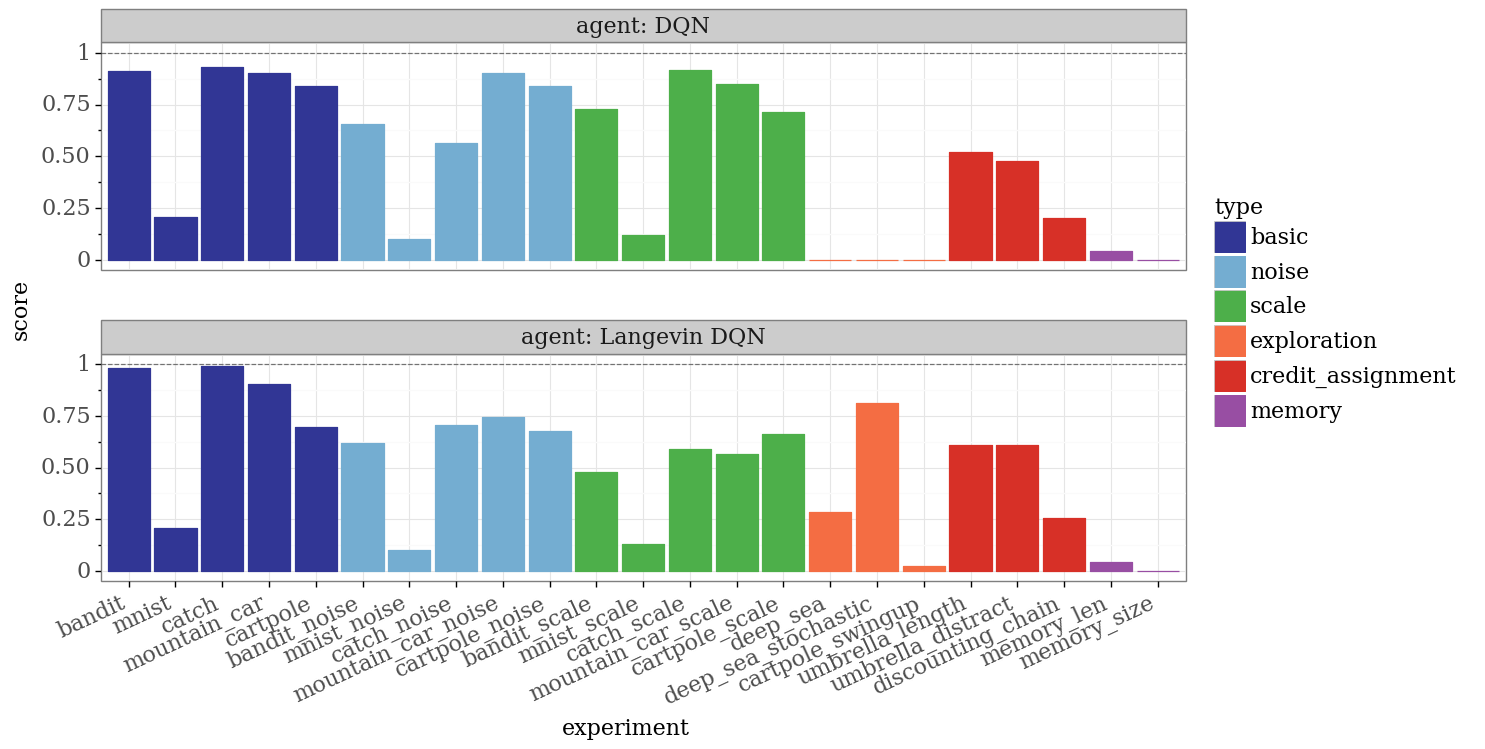}
    \caption{Performance of Langevin DQN and DQN on different bsuite environments.}
    \label{fig:bar-dqn_langevin}
\end{figure}

 Figure \ref{fig:radar-dqn_langevin} shows the performance of Langevin DQN and DQN across different categories of bsuite environments. Details on scoring mechanisms and environments can be found in \citep{osband2020behavior}. Both the agents use $5$ update steps per environment step. Figure \ref{fig:bar-dqn_langevin} shows a more detail comparison of Langevin DQN and DQN on different bsuite environments.

We can see that Langevin DQN performs as well as DQN, if not better, on all categories of bsuite environments except for { \it scale} environments. In the {\it scale} environments, the maximum reward of an environment is scaled with a multiplier ranging from $10^{-3}$ to $10^3$. This test for the robustness of an algorithm against reward scaling, using a single hyperparameter settings. This leads to lower performance in Langevin DQN, as we are using the same value of $\sigma_w^2$ for all reward scales.

\subsection{Ensemble Langevin DQN vs Boot DQN} \label{app:boot_langevin}

Boot DQN is a well-known ensemble sampling algorithm  known for its deep exploration capability \citep{lu2017ensemble, osband2017deep}. Algorithm \ref{alg:ensemble} provides a common template for both Ensemble Langevin DQN and Boot DQN. At the start of each episode, one model is picked uniformly at random and the agent acts greedily w.r.t the picked model throughout the episode. 

\begin{figure}[!ht]
    \centering
    \includegraphics[width=0.75\columnwidth]{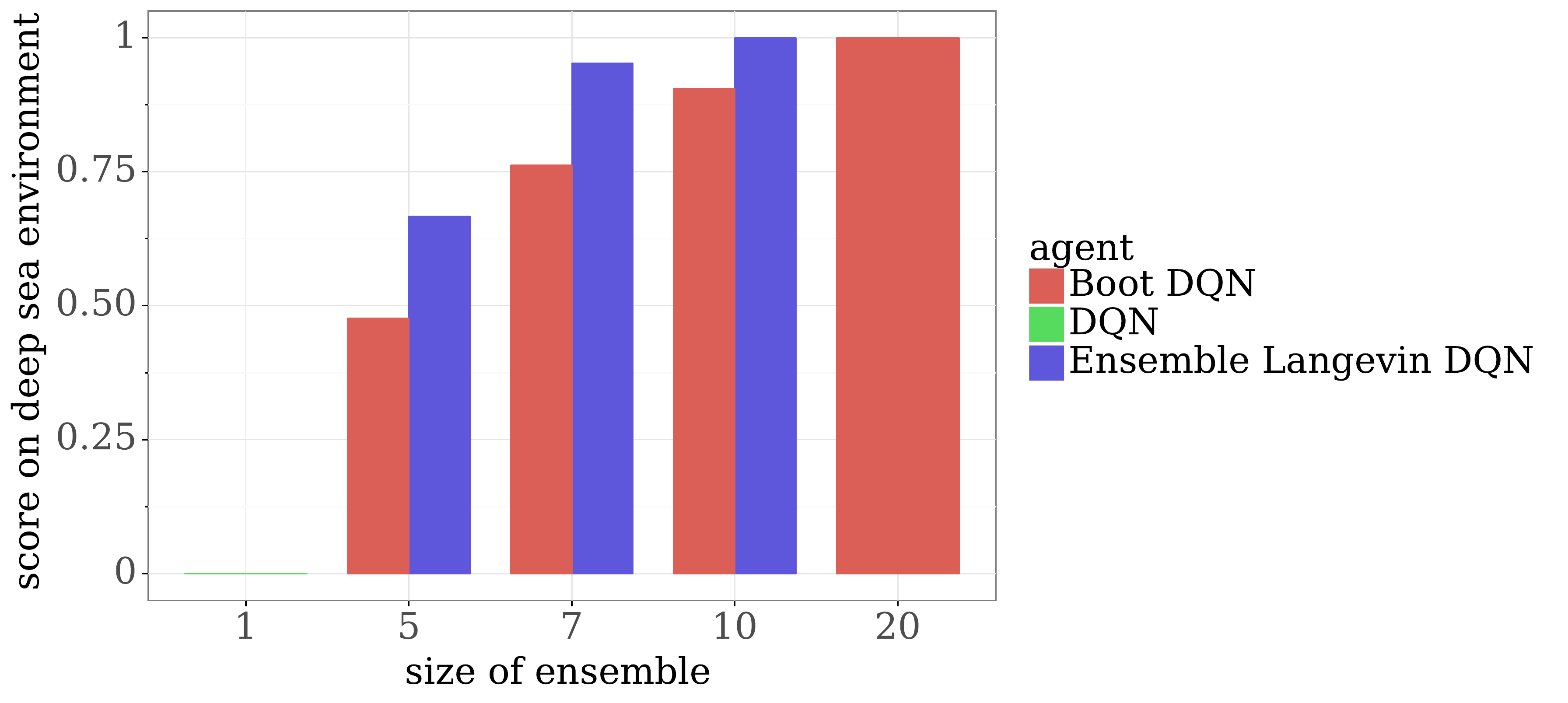}
    \caption{Effect of ensemble size on the performance of Boot DQN and Langevin boot DQN on deep sea environment. Score on the vertical axis is the fraction of deep sea environments in Bsuite that the agent has solved. All the agents use one update step per environment step per model.}
    \label{fig:deep_sea_ensembles}
\end{figure}

Figure \ref{fig:deep_sea_ensembles} shows the effect of ensemble size on the performance of Ensemble Langevin DQN and Boot DQN. All the agents use a single update step per environment step per model for all agents. The horizontal axis shows the number of models in the ensemble and vertical axis shows the score on the deep sea environment, which is the fraction of deep sea environments in the Bsuite that the agent has solved within $10,000$ episodes. Bsuite has $21$ deep sea environments with depth varying from $10$ to $51$ in steps of $2$. Note that the difficulty to solve a deep sea environment increases exponentially with an increase in the depth of the deep sea.

From Figure \ref{fig:deep_sea_ensembles}, we can see that for all ensemble sizes, Ensemble Langevin DQN performs better than Boot DQN. In order to generate this plot,  we use the same hyper parameter setting for all ensemble sizes. We use the parameters mentioned in Appendix \ref{app:param_boot_dqn_langevin} for Boot DQN and Ensemble Langevin DQN, and use the parameters from Appendix \ref{app:param_dqn_langevin} for DQN. We found that the performance of all agents is robust to the changes in the hyperparameters (like $\epsilon$, $\sigma_w^2$) used for generating this plot. We have previously compared the performance of Boot DQN and Ensemble Langevin DQN across a different number of update iterations in Figure \ref{fig:deep_sea_num_sgd}. We can see that the Ensemble Langevin DQN with size $5$ can get a similar performance as a Boot DQN of size $20$ on increasing the number of update iterations, while the performance of Boot DQN remains unchanged with  the number of update iterations.

\begin{figure}[!ht]
    \centering
    \includegraphics[width=0.9\columnwidth]{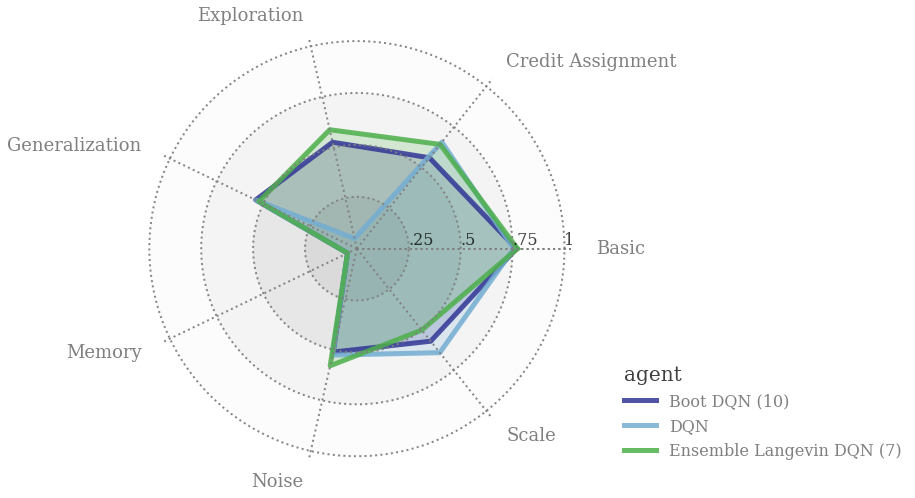}
    \caption{Performance of DQN, Boot DQN, and Ensemble Langevin DQN on different categories of Bsuite environments.}
    \label{fig:bsuite_radar}
\end{figure}
\begin{figure}[!ht]
    \centering
    \includegraphics[width=0.9\columnwidth]{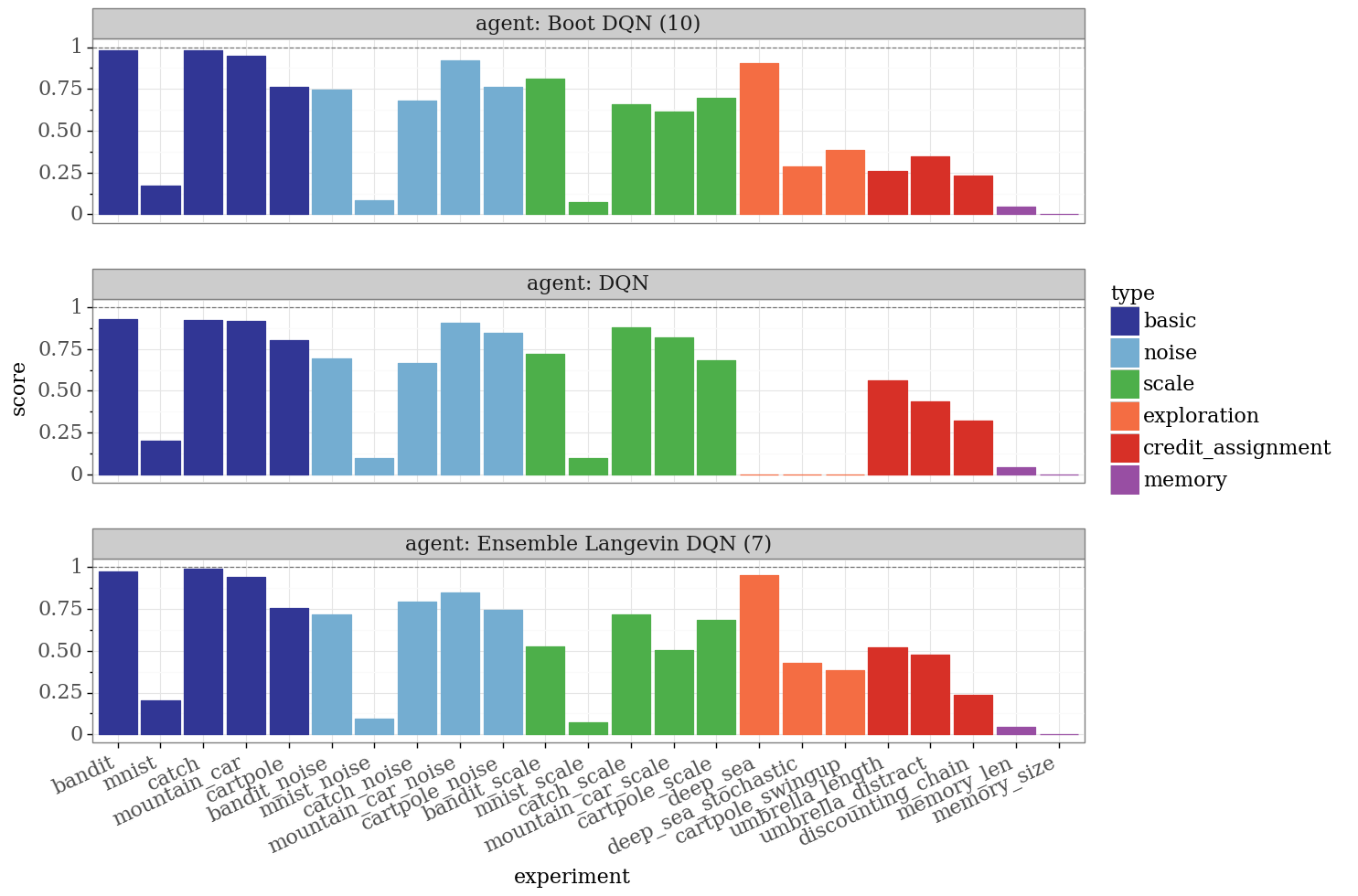}
    \caption{Performance of DQN, Boot DQN, and Ensemble Langevin DQN on different Bsuite environments.}
    \label{fig:bsuite_bar}
\end{figure}

Figure \ref{fig:bsuite_radar} compares the performance of DQN, Boot DQN and Ensemble Langevin DQN across different categories of Bsuite environments. For this experiment, we compare DQN, Boot DQN with an ensemble of size $10$, and Ensemble Langevin DQN with an ensemble of size $7$. From the plot, we can see that both Ensemble Langevin DQN and Boot DQN out perform DQN in terms of their ability to do exploration. In all dimensions except for scale, Ensemble Langevin DQN performs as good as Boot DQN, if not better, even though Ensemble Langevin DQN uses a smaller ensemble than Boot DQN. Figure \ref{fig:bsuite_bar} shows a more detail comparison of agents on different Bsuite environments.

The results presented in the appendix further strengthen our claim that Langevin DQN and Ensemble Langevin DQN are capable of deep exploration. We have also shown that, even though Langevin DQN is a slight variation of DQN, it is capable of deep exploration using a single point estimate without hurting the other capabilities of DQN.

\end{document}